\documentclass[11pt,a4paper]{article}
\usepackage[hyperref]{emnlp2020}
\usepackage{times}
\usepackage{latexsym}

\pdfoutput=1

\usepackage{subfigure}
\usepackage{amsfonts} 
\usepackage{graphicx}
\usepackage{tabularx}
\usepackage{bigstrut}
\usepackage{multirow}
\usepackage{color}

\usepackage{microtype}

\aclfinalcopy % Uncomment this line for the final submission
\title{DocStruct: A Multimodal Method to Extract Hierarchy Structure in Document for General Form Understanding}

\author{Zilong Wang \quad  Mingjie Zhan \quad Xuebo Liu \quad Ding Liang \\ 
        Sensetime \\
        \texttt{\{wangzilong, zhanmingjie, liuxuebo, liangding\}@sensetime.com}}

\date{}

\begin{document}

\maketitle

\begin{abstract}
Form understanding depends on both textual contents and organizational structure. Although modern OCR performs well, it is still challenging to realize general form understanding because forms are commonly used and of various formats. The table detection and handcrafted features in previous works cannot apply to all forms because of their requirements on formats. Therefore, we concentrate on the most elementary components, the key-value pairs, and adopt multimodal methods to extract features. We consider the form structure as a tree-like or graph-like hierarchy of text fragments. The parent-child relation corresponds to the key-value pairs in forms. We utilize the state-of-the-art models and design targeted extraction modules to extract multimodal features from semantic contents, layout information, and visual images. A hybrid fusion method of concatenation and feature shifting is designed to fuse the heterogeneous features and provide an informative joint representation. We adopt an asymmetric algorithm and negative sampling in our model as well. We validate our method on two benchmarks, MedForm and FUNSD, and extensive experiments demonstrate the effectiveness of our method.
\end{abstract}

\section{Introduction}

Forms are a ubiquitous document format. Numerous forms are used in finance, insurance and medical industry every day. Although forms vary a lot, we consider it as a collection of key-value pairs and all these pairs establish a hierarchical structure within the page. Our work in this paper focuses on utilizing multimodal information to extract the hierarchy from the forms. Equipped with the hierarchy, it is oversimplified to further analyze the general forms and extract the structural data.

% the extracted structural information is helpful in understanding and using these data.

% Textual documents are a common source of natural language materials. Confronted with overwhelming number of documents everyday, there is an urgent need to enable computer to process them for us. The first step of automatic processing is to understand the contents of the numerous documents, including the exact meaning as well as the organizational structure. Our work in this paper focuses on the latter part and utilizes information from multiple aspects to extract the hierarchy structure from the document.

\begin{figure}[t]
    \centering
    \includegraphics[width=1.\linewidth]{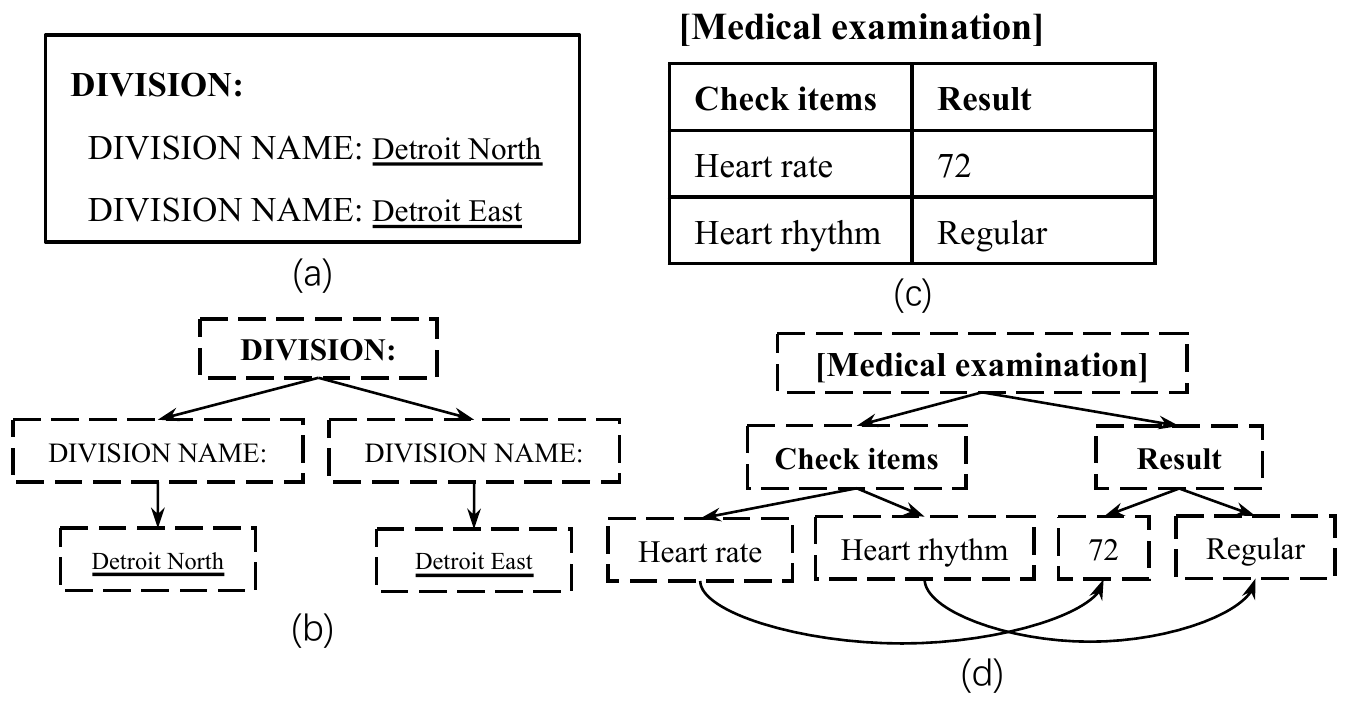}
    \caption{Example from FUNSD Dataset: (a) is part of the original image; (b) is the hierarchical structure. Example from MedForm Dataset: (c) is part of the original image; (d) is the hierarchical structure.}
    \label{fig:example}
\end{figure}

Modern Optical Character Recognition (OCR) has already provided a reliable and efficient way for the computers to read the textual contents of form pages. The contents can be divided into several individual textual fragments. However, it is not enough to form-understanding tasks. 
The information is expressed not only through the textual data in each section, but also through the way in which the sections are organized. 
Some of the fragments serve as headers, topics or questions of their counterparts. We consider the relation as key-value pairs in a hierarchy. Figure \ref{fig:example} are two examples \footnote{The example images are processed and translated in this paper for clearness.
% The original images can be found in Appendix.
}.
Therefore, after preliminary processing with OCR, we need to extract the latent structure in a form page to convert the textual data into structured data. 

% However, it is not enough for some unique types of documents. Many of the documents are composed of form-like contents, i.e., the contents can be divided into several individual sections. Each section is a text fragment which can be an individual phrase, an integrated sentence or a short paragraph. The information in these documents is expressed not only through the textual data in each individual section, but also through the way in which the sections are organized. Therefore, after the preliminary processing with OCR, we need to extract the latent structure in a form-like document page so as to convert the textual data into structural data. 

% To be more specific, there is not necessarily a strict table in the document page, but some asymmetric pairs are enough. 

% In details, a form page consist of several textual fragments. Some of fragments serve as \textit{headers}, \textit{topics} or \textit{questions} of their counterparts. We consider the relation as key-value pairs in a hierarchy. For example, in Figure \ref{fig:example}, although there is no table lines or other key components of forms, we can find several pairs. If we can extract all these pairs, we can build a hierarchy based on them for the document page.

% Hierarchy in human language is ubiquitous and has been studied for a long time. Many previous research focuses on the hierarchical structure among fine-grained semantic units, such as words, phrases. These previous jobs have proposed many innovative methods to deal with the meaning scope of corresponding entities but they ignore the hierarchical relation within the form-like documents pages, which is vital for automatic processing of document. 

% In this paper, we focus on the latent hierarchy in the form-like documents and propose a pipeline which extracts the asymmetric pairs to build the hierarchy. 

% Unlike other form understanding methods which mostly rely on the strict table format, our method focuses on the abstract hierarchical structure. The asymmetric pairs are the simplest components and independent of the any strict formats. Therefore, our method can be applied on a large variety of form-like documents. 

Some related works build their models on the handcrafted features \cite{ha1995recursive,simon1997fast,ha1995document}. They propose heuristic methods and use top-down or bottom-up techniques to build their model. However, the results only reveal coarse structure such as layouts or bounding boxes, which is not enough for actual usage.
% but the results only reveal some coarse structural information such as layouts. 
Others propose table detection techniques \cite{hao2016table,he2017multi}. However, tables are a subset of forms. Table detection doesn't work when there is no table lines or cells in the given form page. Unlike these previous works, our method consider the structure as a hierarchy where parent-child relation corresponds to the key-value pairs. The key-value pairs are the most elementary components and independent of any formats. This ensures a wide range of application of our method and should be the promising direction for general form understanding. 

To comprehensively acquire the informative representation for each fragment and catch the reliable signals for the hierarchical structure, we leverage multimodal information from semantic, layout and visual aspects. We carefully design targeted extraction module for each modality. Following many previous studies, we adopt the pretrained language model, e.g. BERT \cite{devlin2018bert}, RoBERTa \cite{liu2019roberta}, to extract semantic features. The layout and visual information are also useful. For example, nearer fragments should be more likely to be related and fragments with bold faces are more likely to be the title. Therefore, we use multilayer perceptron and character detection algorithms \cite{tian2016detecting} to extract layout and visual features.
To fuse multimodal features, we are enlightened by \citet{wang2019words,rahman2019m} and propose a hybrid fusion method of concatenation and feature shifting. The features differ in meaning and dimension. We utilize the most informative features, semantic and layout features, through concatenation. Then we take vision features as shifting feature to refine fused features. Finally, we design an asymmetric relation prediction module and negative sampling to finish the whole pipeline.

We validate our method, \textit{DocStruct}, on two benchmarks, MedForm and FUNSD \cite{jaume2019funsd}. The first one is built by us and composed of medical examination reports, and the second is composed of various real, fully annotated, scanned forms. We summarize our contribution as follows:
% Our contributions can be summarized as follows:
\begin{itemize}
    \item We focus on the essential components, key-value pairs, and build a hierarchical structure to realize general form understanding, which ensures a large range application of our method.
    \item We adopt a multimodal method and propose a hybrid fusion algorithm to build the form hierarchy from semantic, layout, and vision.
    \item Extensive experiments have been conducted to demonstrate the effectiveness of our method.
\end{itemize}

\section{Methodology}
In this section, we will first describe the preliminary processing step and introduce an overview of our method. Then we propose the \textit{DocStruct} model which extracts and fuses multimodal features and predicts the hierarchical relation between text fragments. Finally, the Negative Sampling training method is also introduced in this section.

\subsection{Overview}
Given a general form page, the text fragments in this page have been extracted by Optical Character Recognition (OCR) or by human labors beforehand. Each fragment contains complete semantic meaning (e.g., an individual phrase, an integrated sentence, a short paragraph). We aim at building the latent hierarchical structure of these fragments in the page. 

The key-value relation between the extracted text fragments depends on multiple aspects. The preliminary processing provides us with semantic contents and layout information of each fragment. We further crop the image segments from the original page for visual information. We want to extract multimodal features from the semantic, layout and visual information and fuse them through a carefully-designed algorithm with regard to their differences. Equipped with the informative joint feature, we predict the superior counterpart of each fragment. As long as each fragment finds its corresponding superior counterpart, we can construct a tree-like or graph-like hierarchy accordingly.

We denote a form page as $D$ and a fragment as $X$. The page $D$ is represented as a set of fragments: $D = [X_1, X_2,...,X_n]$, where $n$ is the number of fragments in this page. We have three multimodal features for an individual fragment, so we denote them as $X_i = (X^{S}_i, X^{L}_i, X^{V}_i)$ for semantic, layout, visual features, respectively. The hierarchical structure of the fragments is represented as directed edges between fragments. We denote the edges as $X_i \rightarrow X_j$, which means fragment $i$ serves as a topic or header of fragment $j$ and should be considered superior to fragment $j$.

\begin{figure*}[t]
    \centering
    \includegraphics[width=\linewidth]{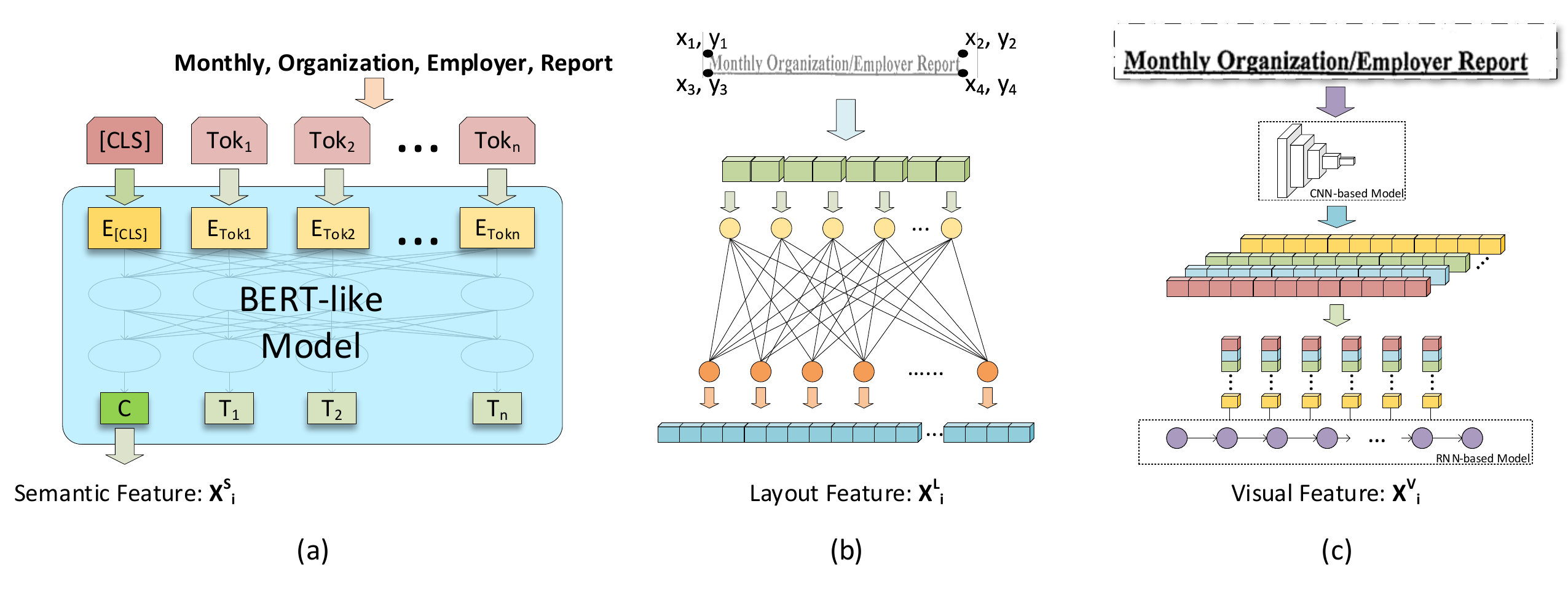}
    \caption{Feature Extraction Modules: (a), (b), (c) are for semantic, layout, visual features, respectively.}
    \label{fig:featEx}
\end{figure*}

\subsection{Proposed Model}
Our proposed model, \textit{DocStruct}, contains: three targeted feature extraction modules for semantic, layout, and visual information, respectively, a feature fusion module and a relation prediction module.

\subsubsection{Semantic Feature Extraction Module}
The semantic content of each text fragment is acquired from the results of OCR or human labors. Intuitively, semantic contents are reliable signals to predict the hierarchical relation. We follow many previous works and use BERT-like pretrained language models to extract semantic features for each text fragment. Numerous natural language processing tasks have demonstrated the outstanding performance of pretrained models' ability to extract textual features. These models are designed to give the deep bidirectional representations from extensive unlabeled corpus with regard to both left and right context. More importantly, independent of the large corpus in pretraining step, the outputs of these models can be easily used for downstream tasks. A special tag \verb|[CLS]| is added in front of the inputs and the corresponding output can be used for fine-tuning.

We first input the raw semantic contents to a BERT-like pretrained language model and select the \verb|[CLS]|'s hidden state of the last layer as the semantic feature for the text fragment.
\begin{equation}
    T_i = [[CLS], w_1, w_2,..., w_n]
\end{equation}
\begin{equation}
    H_i = Bert(T_i)
\end{equation}
\begin{equation}
    X^S_i = H_i[0] \in \mathbb{R}^{d^S}
\end{equation}
where $T_i$ is the raw contents of text fragment $i$ with the added tag \verb|[CLS]|; $Bert$ is a symbol for BERT-like pretrained models; $H_i$ is the last hidden states of the pretrained model; and $X^S_i$ is our extracted semantic feature for the fragment $i$. The dimension of $X^S_i$ equals to the hidden states of pretrained model, and we denote it as $d^S$.

\subsubsection{Layout Feature Extraction Module}
The preliminary processing of OCR or human labor also offers the layout information: the relative coordinates of the text fragment's vertices. The layout information shows the size and relative location of the text fragment, which helps to distinguish different text fragments with the same semantic contents. 

We calculate the rectangular closure with the coordinates and input the 8-dimension vector into a fully connected layer to project the vector into a hyperspace.
\begin{equation}
    C_i = [x_1, y_1, x_2, y_2, x_3, y_3, x_4, y_4]
\end{equation}
\begin{equation}
    X^L_i = \sigma(WC_i+b) \in \mathbb{R}^{d^L}
\end{equation}
where $C_i$ is the coordinates of rectangular closure's vertices; $W$ is the weight matrix; $b$ is the bias; $\sigma(\cdot)$, the activation function, is set as $relu(\cdot)$; and $X^L_i$ is the layout feature for the fragment $i$. The dimension of $X^L_i$ is a hyper-parameter, and we denote it as $d^L$.

% visual feature extraction part needs revision! the parts about text detection need more details!
\subsubsection{Visual Feature Extraction Module}
Visual information is the image part cropped from the original page with the rectangular closure of the fragment. Intuitively, visual information also provides worthwhile signals to predict the key-value relation. For example, bold faces are more likely to be superior.

Since the concerning images are parts from pages, they mostly consist of letters or characters and are unlikely to include ordinary pictures. This makes some generic methods (e.g., Resnet, VGG) unsuitable in this situation. We notice that the features we care about mostly concentrate on the style of characters, such as the bold faces, italics, etc. Enlightened by text detection tasks, we use a deep CNN-based model to extract a feature map, followed by an RNN-based model considering the textual sequence. 

The CNN-based model is carefully designed. The height of the output feature map is 1 and we concatenate the features of each channel. If we view the concatenated features along the width of the feature map, they are a new sequential inputs for the RNN-based model. Each time step of the RNN-based model symbolizes a frame of the image, corresponding to each letter. Finally, Max Pooling is used to extract the most significant features from the hidden states of RNN.
\begin{equation}
    M_i = CNN(I_i) \in \mathbb{R}^{c,h,w} \ \ \ (h=1)
\end{equation}
\begin{equation}
    M^{\prime}_i = trans(M_i) \in \mathbb{R}^{w,c}
\end{equation}
\begin{equation}
    F_i = RNN(M^{\prime}_i)
\end{equation}
\begin{equation}
    X^V_i = Max(F_i) \in \mathbb{R}^{d^V}
\end{equation}
where $I_i$ is the image part of the fragment $i$; $CNN$, the CNN-based model, is set as Resnet50 with minor changes on the last pooling layer to fit the height restriction; $c,h,w$ are the channel, height, width of the feature map, respectively; the $trans(\cdot)$ is to convert the original feature map into a sequential input; $M^{\prime}_i$; $RNN$, the RNN-based model, is set as a two-layer bi-directional LSTM; $Max(\cdot)$ is the function of Max Pooling, conducted on the hidden states $F_i$; and $X^V_i$ is the visual features for fragment $i$. The dimension of $X^V_i$ is denoted as $d^V$ and set as $d^S+d^L$ to fit the requirement of Feature Fusion Module.

\begin{figure}[h]
    \centering
    \includegraphics[width=0.8\linewidth]{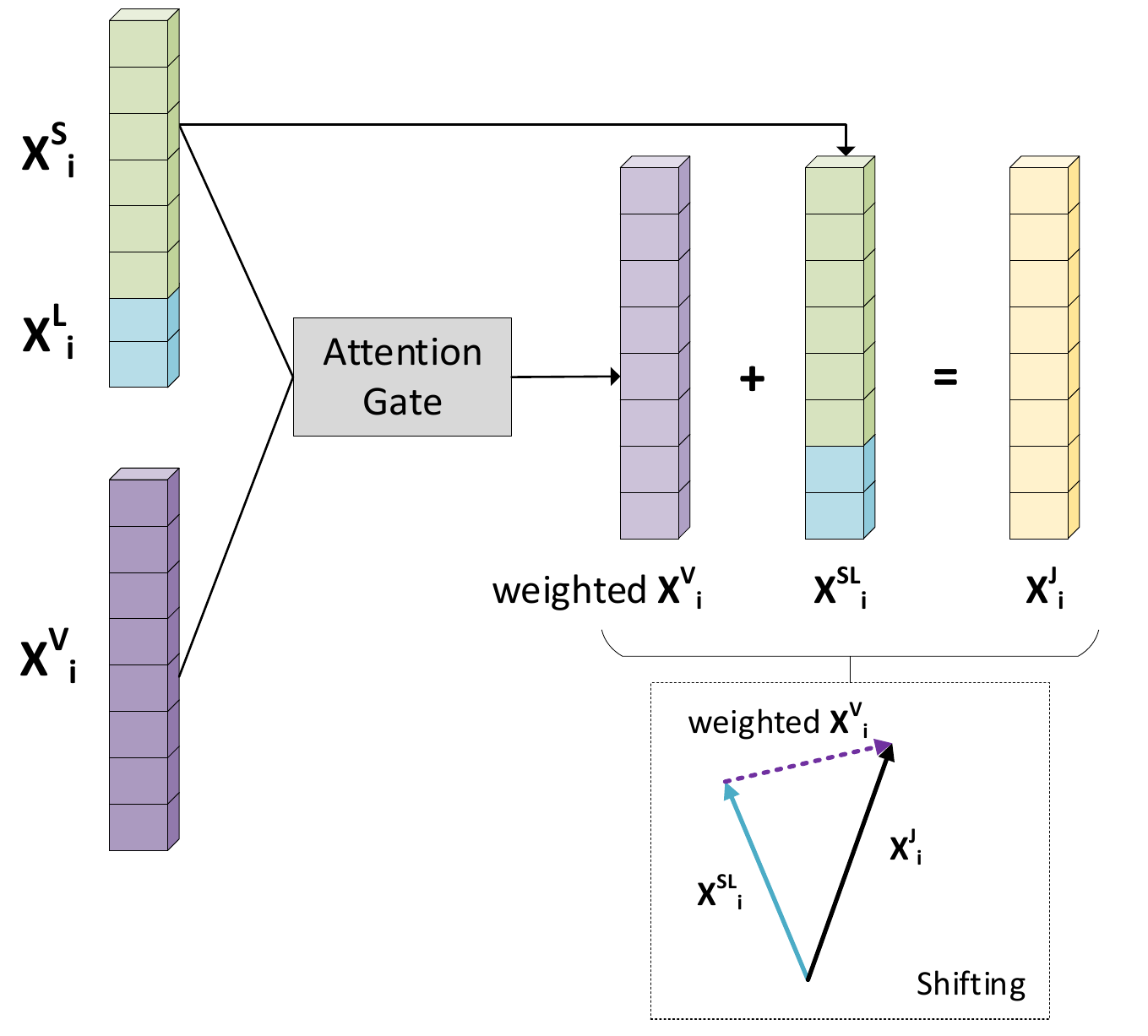}
    \caption{Feature Fusion Module: $X^S_i$, $X^L_i$, $X^V_i$ are semantic, layout, and visual features respectively. The attention gate takes all features to calculate weights for $X^V_i$. The weighted $X^V_i$ is the shifting feature.}
    \label{fig:fusion}
\end{figure}

\subsubsection{Feature Fusion Module}
% ConcatShiftAddFusion
With the feature extraction modules, we extract multimodal features from semantic, layout and vision aspects. We design a hybrid fusion algorithm to leverage the heterogeneous features of different dimensions and meanings. 
There are two major challenges:

\paragraph{Different dimension:}
The features are of different dimensions, which makes them have different significance in fusion calculation. Since the layout feature is projected from a very low-dimension coordinate feature (only 8-dimension). That makes $d^L$ much smaller than $d^S$ and $d^V$ which are from much more sources, semantic contents and images. 

\paragraph{Different meaning:}
The features from different modalities have different meanings. Intuitively, they contribute unevenly to the final prediction. Semantic features and layout feature should be the most reliable one and the layout feature can also distinguish the fragments with the same contents. The visual feature also provide some additional style signals and help to cope with some problematic cases.

Accordingly, we should not treat the multimodal features in the same way and have to consider their differences.

Inspired by the multi-modality fusion tasks in \citet{poria2017context,wang2019words}, we propose a hybrid fusion algorithm with regard to the differences. We first follow the most direct fusion method, concatenation, and concatenate the semantic feature and layout feature as the base feature. According to the experiments, this semantic-layout feature already performs well in prediction task and there is a considerable increment compared with the individual feature. Then we leverage visual feature to fix minor mistakes in prediction but do not want the extra features to influence the satisfactory results. Following \citet{wang2019words}, the visual feature is used as the shifting feature. We adopt an attention-based influence gate to control the influence from visual feature. To be more specific, we use a fully connected layer on the concatenation of three features to calculate the required weight. Finally, we add the weighted visual feature to the semantic-layout feature and obtain the joint representation considering all the three modalities.
\begin{equation}
    X^{SL}_i = [X^S_i;X^L_i]
\end{equation}
\begin{equation}
    \alpha_i = \sigma(W[X^S_i;X^L_i;X^V_i]+b) \in \mathbb{R}
\end{equation}
\begin{equation}
    X^{J}_i = X^{SL}_i + \alpha_i X^V_i \in \mathbb{R}^{d^V} = \mathbb{R}^{d^S+d^L}
\end{equation}
where $[;]$ is a concatenation of corresponding vectors; $X^{SL}_i$ is the semantic-layout feature; $\sigma(\cdot)$, the activation function, is set as $sigmoid(\cdot)$ here; $W$ and $b$ are the weight matrix and bias; $\alpha_i$ is the gate weight; $X^J_i$ is the joint feature of three modalities.

\subsubsection{Relation Prediction Module}
This module is designed to predict the relation between two given text fragments and use a scalar to evaluate the probability of a directed edge between them. 
The hierarchical relation we care about is asymmetric, i.e., the probability of $X_i \rightarrow X_j$ is completely different from that of $X_j \rightarrow X_i$. Therefore, some common symmetric methods, e.g. Dot Production, Euclidean distance, Poincar\'e distance \cite{nickel2017poincare}, are no longer feasible. They can be proper metric for correlation but cannot further evaluate the probability for asymmetric pairs.

To model the asymmetric relation, the function $\mathcal{P}_{i\rightarrow j}$ needs to meet the requirement that: $\mathcal{P}_{i\rightarrow j} \not= \mathcal{P}_{j\rightarrow i}$.
We utilize a parameter matrix to model the asymmetric relation. We put the matrix in the middle of two joint features. Given the joint features of two text fragments, $X^J_i$ and $X^J_j$, the probability of $X_i \rightarrow X_j$ is calculated through:
% We design $\mathcal{P}_{i\rightarrow j}$ as followed:
\begin{equation}
    \mathcal{P}_{i\rightarrow j} = X^J_j M (X^J_i)^T \in \mathbb{R} \label{eq:relation}
\end{equation}
where $M$ is an asymmetric parameter matrix, so $\mathcal{P}_{i\rightarrow j} \not= \mathcal{P}_{j\rightarrow i}$. 
% Furthermore, 
% \begin{equation}
%     \frac{\partial \mathcal{P}_{i\rightarrow j}}{\partial M} = X^J_j (X^J_i)^T; \frac{\partial \mathcal{P}_{j\rightarrow i}}{\partial M} = X^J_i (X^J_j)^T
% \end{equation}
% Because $X^J_j (X^J_i)^T \cdot X^J_i (X^J_j)^T = ||X^J_i||_2||X^J_j||_2$, the gradient of $\mathcal{P}_{j\rightarrow i}$ grows when gradient of $\mathcal{P}_{i\rightarrow j}$ is reduced. In this way, we build the relation prediction module meeting the requirements.

% $\frac{\partial \mathcal{P}_{i\rightarrow j}}{\partial M} = X^J_j (X^J_i)^T$ and $\frac{\partial \mathcal{P}_{j\rightarrow i}}{\partial M} = X^J_i (X^J_j)^T$. Because $X^J_j (X^J_i)^T \cdot X^J_i (X^J_j)^T = ||X^J_i||_2||X^J_j||_2$. The designed function meets the second requirement.

% We utilize a parameter matrix to model the asymmetric relation. We put the matrix in the middle of two joint features. Given the joint features of two text fragment, $X^J_i$ and $X^J_j$, the probability of $X_i \rightarrow X_j$ is calculated through:

\subsection{Training Method}
It should be noted that the directed edges of the hierarchy only exist between some pairs of text fragments. There may be a key-value relation between two text fragments, but it is more likely that the two random selected fragments are not related at all. To handle the data sparsity and balance the different number of related and unrelated pairs, we adopt the Negative Sampling method \cite{mikolov2013distributed} to train our model. 

Given an asymmetric pair, $X_i \rightarrow X_j$, which means fragment $i$ and $j$ are related and $i$ is superior to $j$, we random select a fix number of unrelated or inferior counterparts for fragment $j$. We generate a negative sampling set of fixed size for $X_j$, i.e., $X_k \not\rightarrow X_j$ for any fragment $k$ in the negative sampling set. 

We choose to build a negative sampling set of the superior side but not the inferior side because one superior fragment may correspond to many inferior ones, but one inferior fragment may only correspond to one or two superior counterparts. The hierarchy in the document page is more like a tree than a graph.

As for fragment $j$, the training target is to enable our model to distinguish the $X_i$ from the negative samples, $X_k$. We normalize the probability and minimize the cross entropy of $\mathcal{P}_{i\rightarrow j}$. So we can maximize the $\mathcal{P}_{i\rightarrow j}$ and minimize the $\mathcal{P}_{k\rightarrow j}$.
\begin{equation}
    \mathcal{L} = -\log \sum_{i \rightarrow j} \frac{e^{\mathcal{P}_{i\rightarrow j}}}{e^{\mathcal{P}_{i\rightarrow j}}+\sum_{k \in Neg(j)} e^{\mathcal{P}_{k \rightarrow j}}}
\end{equation}
where $Neg(j)$ is the negative sampling set of fragment $j$ and $\mathcal{P}_{i\rightarrow j}$ is the probability of a directed edge existing between fragment $i$ and $j$.

% \begin{table*}[t]
%   \centering
%   \caption{Ablation Study Results}
% \begin{tabular}{|c|cc|ccc|cc|ccc|}
% \hline
%       & \multicolumn{5}{c|}{MedForm}          & \multicolumn{5}{c|}{FUNSD} \bigstrut\\
% \cline{2-11}      & \multicolumn{2}{c|}{Reconstruction} & \multicolumn{3}{c|}{Detection} & \multicolumn{2}{c|}{Reconstruction} & \multicolumn{3}{c|}{Detection} \bigstrut\\
% \cline{2-11}Features & mAP   & mRank & Hit@1 & Hit@2 & Hit@5 & mAP   & mRank & Hit@1 & Hit@2 & Hit@5 \bigstrut\\
% \hline
% \hline
% DocStruct(S)     & 0.5928 & 4.48  & 56.96 & 76.24 & 91.85 & 0.4498 & 8.61  & 31.27 & 45.57 & 65.51 \bigstrut[t]\\
% DocStruct(L)     & 0.5085 & 7.07  & 38.23 & 55.96 & 78.18 & 0.6295 & 3.75  & 48.35 & 64.17 & 82.79 \\
% DocStruct(V)     & 0.2744 & 22.56 & 17.24 & 27.83 & 44.20 & 0.2145 & 14.99 & 9.62  & 15.91 & 29.52 \bigstrut[b]\\
% \hline
% FUNSD-base  & 0.3019 & 13.33  & 15.79 & 27.98 & 49.52 & 0.2385 & 11.68  & 10.12 & 16.26 & 36.20 \bigstrut\\
% LayoutLM  & - & -  & - & - & - & 0.4761 & 7.11  & 32.43 & 45.56 & 66.41 \bigstrut\\
% % \hline
% DocStruct(S, L)  & 0.8641 & 2.10  & 84.85 & 92.68 & 97.14 & 0.7043 & 2.96  & 55.94 & 75.48 & 88.46 \bigstrut\\
% \hline
% \textbf{DocStruct} & \textbf{0.8903} & \textbf{1.85} & \textbf{88.41} & \textbf{94.63} & \textbf{98.07} & \textbf{0.7177} & \textbf{2.89} & \textbf{58.19} & \textbf{76.27} & \textbf{88.94} \bigstrut\\
% \hline
% \end{tabular}%
%   \label{tab:ablation}%
% \end{table*}%

% Table generated by Excel2LaTeX from sheet 'Sheet2'
\begin{table*}[t]
    \centering
  \caption{Ablation Study Results}
      \begin{tabular}{|c|c|c|c|c|c|}
      \hline
            & \multicolumn{5}{c|}{MedForm} \bigstrut\\
  \cline{2-6}          & \multicolumn{2}{c|}{Reconstruction} & \multicolumn{3}{c|}{Detection} \bigstrut\\
  \cline{2-6}    Features & \multicolumn{1}{c}{mAP} & mRank & \multicolumn{1}{c}{Hit@1} & \multicolumn{1}{c}{Hit@2} & Hit@5 \bigstrut\\
      \hline
      \hline
      DocStruct(S) & 0.5164 & 5.79  & 47.41 & 66.75 & 85.91 \bigstrut[t]\\
      DocStruct(L) & 0.5085 & 7.07  & 38.23 & 55.96 & 78.18 \\
      DocStruct(V) & 0.2744 & 22.56 & 17.24 & 27.83 & 44.20 \bigstrut[b]\\
      \hline
      FUNSD-base & 0.3019 & 13.33 & 15.79 & 27.98 & 49.52 \bigstrut[t]\\
      LayoutLM & -     & -     & -     & -     & - \\
      DocStruct(S,L) & 0.8641 & 2.10  & 84.85 & 92.68 & 97.14 \bigstrut[b]\\
      \hline
      \textbf{DocStruct} & \textbf{0.8903} & \textbf{1.85} & \textbf{88.41} & \textbf{94.63} & \textbf{98.07} \bigstrut\\
      \hline
      \multicolumn{1}{c}{} & \multicolumn{1}{c}{} & \multicolumn{1}{c}{} & \multicolumn{1}{c}{} & \multicolumn{1}{c}{} & \multicolumn{1}{c}{} \bigstrut\\
      \hline
            & \multicolumn{5}{c|}{FUNSD} \bigstrut\\
  \cline{2-6}          & \multicolumn{2}{c|}{Reconstruction} & \multicolumn{3}{c|}{Detection} \bigstrut\\
  \cline{2-6}    Features & \multicolumn{1}{c}{mAP} & mRank & \multicolumn{1}{c}{Hit@1} & \multicolumn{1}{c}{Hit@2} & Hit@5 \bigstrut\\
      \hline
      \hline
      DocStruct(S) & 0.4498 & 8.61  & 31.27 & 45.57 & 65.51 \bigstrut[t]\\
      DocStruct(L) & 0.6295 & 3.75  & 48.35 & 64.17 & 82.79 \\
      DocStruct(V) & 0.2145 & 14.99 & 9.62  & 15.91 & 29.52 \bigstrut[b]\\
      \hline
      FUNSD-base & 0.2385 & 11.68 & 10.12 & 16.26 & 36.20 \bigstrut[t]\\
      LayoutLM & 0.4761 & 7.11  & 32.43 & 45.56 & 66.41 \\
      DocStruct(S,L) & 0.7043 & 2.96  & 55.94 & 75.48 & 88.46 \bigstrut[b]\\
      \hline
      \textbf{DocStruct} & \textbf{0.7177} & \textbf{2.89} & \textbf{58.19} & \textbf{76.27} & \textbf{88.94} \bigstrut\\
      \hline
      \end{tabular}%
  \label{tab:ablation}%
  \end{table*}%

% [eval: 0.3019 mAP: 0.3019 mRank: 13.3301 top1: 0.1579 top2: 0.2798 top5: 0.4952] best [0.3019]
\section{Experiment}
In this section, we conduct experiments on two benchmarks, MedForm and FUNSD \cite{jaume2019funsd} \footnote{\url{https://guillaumejaume.github.io/FUNSD/}}, to validate the effectiveness of our proposed model for building the latent hierarchy in forms. We design two tasks, \textit{Reconstruction} and \textit{Detection}. The metric explanation is in Appendix.

\begin{table}[htbp]
  \centering
  \caption{Statistics of the Datasets}
%   \resizebox{\linewidth}{14mm}{
\begin{tabular}{|c|cccc|}
\hline
      & Split & Pages & Frag. & Pairs \bigstrut\\
\hline
\hline
\multirow{2}[2]{*}{MedForm} & Train & 686   & 53444     & 44976 \bigstrut[t]\\
      & Test  & 171   & 14013   & 12281  \bigstrut[b]\\
\hline
\multirow{2}[2]{*}{FUNSD} & Train & 149   & 7411  & 4236 \bigstrut[t]\\
      & Test  & 50    & 2332  & 1076 \bigstrut[b]\\
\hline
\end{tabular}%

    % }
  \label{tab:dataset}%
\end{table}%

\paragraph{Reconstruction:} Given the labeled hierarchical structure in a document, we predict the superior counterpart for each text fragment so as to rebuild the hierarchy. the Mean Average Precision (mAP) and Mean Rank (mRank) are used as metrics.
\paragraph{Detection:} To test the detection ability of our model, we choose the counterpart with the highest probability as the prediction result and calculate the Hit@1, Hit@2 and Hit@5 as metrics.

\subsection{Datasets}
We select two datasets, MedForm and FUNSD. The statistics of these two datasets are listed in Table \ref{tab:dataset}. More detailed descriptions are in Appendix.

% \paragraph{MedForm:} We collect a large number of Chinese medical examination reports and build the dataset \textit{MedForm}. The report does not come from one institution, which means that the formats will be different. This adds the difficulty to analyze the structure. We first process the pages through human labors so as to acquire the perfect recognition of textual contents and layout information.

% \paragraph{FUNSD:} We also select the dataset \textit{FUNSD} as our benchmark. \textit{FUNSD} is composed of 199 real, fully annotated, scanned forms. The documents are noisy and vary widely in appearance, making form understanding a challenging task. This dataset labels the position of single words and the links between text fragments.

\begin{itemize}
    \item \textbf{MedForm:} We collect a large number of Chinese medical examination reports and build the dataset \textit{MedForm}. The report does not come from one institution, which means that the formats will be different. This adds the difficulty to analyze the structure. We first process the pages through human labors so as to acquire the perfect recognition of textual contents and layout information.
    \item \textbf{FUNSD:} We also select the dataset \textit{FUNSD} as our benchmark. \textit{FUNSD} is composed of 199 real, fully annotated, scanned forms. The documents are noisy and vary widely in appearance, making form understanding a challenging task. This dataset labels the position of single words and the links between text fragments.

\end{itemize}

% \subsection{Training Details}
% In semantic feature extraction module, we use $BERT_{Large}$ in \citet{devlin2018bert} for FUNSD dataset and use $RoBERTa$ in \citet{chinese-bert-wwm} for MedForm dataset because of their difference in language. The dimension of semantic feature $d^S$ is 1024. In layout feature extraction module, we set the feature dimension $d^L$ as 128. In visual feature extraction module, the CNN-based model we use is $Resnet50$ in \citet{he2016deep} and the RNN-based model is a two-layer bidirectional $LSTM$ whose hidden size is $d^V/2$. The size of negative sampling set is set as 50. The batch size is 4 pages and the learning rate is 0.01. We conduct manual tuning. 

\subsection{Ablation Study and Baseline Comparison}
We acquire features from three different modalities, semantic, layout and vision. Ablation study is conducted to demonstrate the effectiveness of each modality. We remove some of the features and construct several comparable baselines. 
We also select two baselines.
\begin{itemize}
    \item \textbf{FUNSD-base:} The baseline is offered in \citet{jaume2019funsd}, which uses semantic and layout information.
    \item \textbf{LayoutLM:} We replace the feature extraction modules with LayoutLM \cite{xu2020layoutlm}. Since it uses the layout information of single words, which is not provided by MedForm, the experiments are not conducted on MedForm. 
\end{itemize}

We compare our proposed method with these baselines to show the improvements. The statistics are listed in Table \ref{tab:ablation}. In the feature column, S, L, V refers to semantic feature, layout feature, and vision features, respectively. In the baseline of \{S, L\}, we calculate the joint representation through the concatenation of semantic features and layout features. The relation prediction module is the same except for the different matrix dimension.

% \begin{table*}[h]
%   \centering
%   \caption{Ablation Study Results}
% \begin{tabular}{|c|cc|ccc|cc|ccc|}
% \hline
%       & \multicolumn{5}{c|}{MedForm}          & \multicolumn{5}{c|}{FUNSD} \bigstrut\\
% \cline{2-11}      & \multicolumn{2}{c|}{Reconstruction} & \multicolumn{3}{c|}{Detection} & \multicolumn{2}{c|}{Reconstruction} & \multicolumn{3}{c|}{Detection} \bigstrut\\
% \cline{2-11}Features & mAP   & mRank & Hit@1 & Hit@2 & Hit@5 & mAP   & mRank & Hit@1 & Hit@2 & Hit@5 \bigstrut\\
% \hline
% \hline
% S     & 0.5928 & 4.48  & 56.96 & 76.24 & 91.85 & 0.4498 & 8.61  & 31.27 & 45.57 & 65.51 \bigstrut[t]\\
% L     & 0.5085 & 7.07  & 38.23 & 55.96 & 78.18 & 0.6295 & 3.75  & 48.35 & 64.17 & 82.79 \\
% V     & 0.2744 & 22.56 & 17.24 & 27.83 & 44.20 & 0.2145 & 14.99 & 9.62  & 15.91 & 29.52 \bigstrut[b]\\
% \hline
% S, L  & 0.8641 & 2.10  & 84.85 & 92.68 & 97.14 & 0.7043 & 2.96  & 55.94 & 75.48 & 88.46 \bigstrut\\
% \hline
% \textbf{Ours} & \textbf{0.8903} & \textbf{1.85} & \textbf{88.41} & \textbf{94.63} & \textbf{98.07} & \textbf{0.7177} & \textbf{2.89} & \textbf{58.19} & \textbf{76.27} & \textbf{88.94} \bigstrut\\
% \hline
% \end{tabular}%
%   \label{tab:ablation}%
% \end{table*}%

From the results, we find that the performance of both tasks on both datasets improves with more modalities considered. If we only consider single modality, semantic features offer strong signals to judge the hierarchical relation but it cannot deal with the fragment with the same textual contents, which is responsible for the corresponding low performance. The layout and visual information can be a great help, but they serve as auxiliary features since they are not as informative as semantic features. A great improvement can be seen if we merge semantic features and layout features together. After adding the visual features, all the metric factors are even higher. This proves the effectiveness of the hybrid feature fusion module.

% Table generated by Excel2LaTeX from sheet 'Sheet3'
\begin{table*}[t]
  \centering
  \caption{Fusion Method Comparison}
  % Table generated by Excel2LaTeX from sheet 'Sheet3'
\begin{tabular}{|c|cc|ccc|}
\hline
      & \multicolumn{5}{c|}{MedForm} \bigstrut\\
\cline{2-6}Fusion method & mAP   & mRank & Hit@1 & Hit@2 & Hit@5 \bigstrut\\
\hline
\hline
Concatenation & 0.8706 & 2.10  & 85.87 & 92.68 & 97.23 \bigstrut[t]\\
Concat. + Feature shifting w/o gate & 0.8677 & 2.07  & 85.23 & 92.92 & 97.34 \bigstrut[b]\\
\hline
\textbf{Concat. + Feature shifting with gate (ours)} & \textbf{0.8903} & \textbf{1.85} & \textbf{88.41} & \textbf{94.63} & \textbf{98.07} \bigstrut\\
\hline
\multicolumn{1}{r}{} &       & \multicolumn{1}{r}{} &       &       & \multicolumn{1}{r}{} \bigstrut\\
\hline
      & \multicolumn{5}{c|}{FUNSD} \bigstrut\\
\cline{2-6}Fusion method & mAP   & mRank & Hit@1 & Hit@2 & Hit@5 \bigstrut\\
\hline
\hline
Concatenation & 0.7028 & 2.97  & 56.36 & 74.38 & 87.71 \bigstrut[t]\\
Concat. + Feature shifting w/o gate & 0.7024 & 3.10  & 55.82 & 74.76 & 88.84 \bigstrut[b]\\
\hline
\textbf{Concat. + Feature shifting with gate (ours)} & \textbf{0.7177} & \textbf{2.89} & \textbf{58.19} & \textbf{76.27} & \textbf{88.94} \bigstrut\\
\hline
\end{tabular}%

  \label{tab:fusion}%
\end{table*}%

\subsection{Fusion Method Comparison}
From the results of ablation study, we observe that multimodal features contribute unevenly to the final prediction. Because of the different dimension and different meaning of features, it is vital to figure out the proper method to leverage all the aspects and fuse them together. 

In our proposed method, we adopt a hybrid feature fusion method of concatenation and feature shifting, which uses gate mechanism to control the weight of the shifting feature to be added to the concatenated feature. In our settings, the visual feature serves as the shifting feature and the base feature is the concatenation of semantic and layout feature. We adopt another two different feature fusion methods as comparison to demonstrate the effectiveness of our adopted method.

\paragraph{Concatenation (Concat.):} \textit{Concatenation} is the simplest fusion method, which concatenates all the concerning features and produces a long vector as joint representation.

\paragraph{Concat. + Feature shifting w/o gate (Concat. Shift w/o gate):} Our proposed hybrid fusion method (\textit{Concat. + Feature shifting with gate, Concat. Shift}) uses an attention gate to calculate weight for vision feature to be added to the base feature. In \textit{Concat. + Feature shifting w/o gate, (Concat. Shift w/o gate)}, the gate mechanism is removed, i.e., the weight is always 1. The joint representation is calculated by the sum of base feature and shifting feature.

The comparison results are shown in Table \ref{tab:fusion}. From the results, we can see the \textit{Concat. + Feature shifting with gate} achieves the highest performance. We also find that adding more features does not always mean better performance. In the experiments on FUNSD dataset, the reconstruction results of \textit{Concatenation} and \textit{Concat. + Feature shifting fusion w/o gate} perform even worse than the comparative baseline of \{S, L\}. This shows that the extra features may interfere with the existing features unless they are fused by a proper method. That's why we propose the hybrid fusion method to merge the concerning features with regard to their differences. We would like to control the influence through the attention-based gate, and the increments demonstrate the effectiveness.

% % Table generated by Excel2LaTeX from sheet 'Sheet3'
% \begin{table*}[t]
%   \centering
%   \caption{Fusion Method Comparison on MedForm}
% % Table generated by Excel2LaTeX from sheet 'Sheet3'
% \begin{tabular}{|c|cc|ccc|}
% \hline
%       & \multicolumn{5}{c|}{MedForm} \bigstrut\\
% \cline{2-6}Fusion method & mAP   & mRank & Hit@1 & Hit@2 & Hit@5 \bigstrut\\
% \hline
% \hline
% Concatenation & 0.8706 & 2.10  & 85.87 & 92.68 & 97.23 \bigstrut[t]\\
% Feature shifting w/o gate & 0.8677 & 2.07  & 85.23 & 92.92 & 97.34 \bigstrut[b]\\
% \hline
% \textbf{Feature shifting with gate (ours)} & \textbf{0.8903} & \textbf{1.85} & \textbf{88.41} & \textbf{94.63} & \textbf{98.07} \bigstrut\\
% \hline
% \end{tabular}%
%   \label{tab:fusion_a}%
% \end{table*}%

% % Table generated by Excel2LaTeX from sheet 'Sheet3'
% \begin{table*}[t]
%   \centering
%   \caption{Fusion Method Comparison on FUNSD}
%     % Table generated by Excel2LaTeX from sheet 'Sheet3'
% \begin{tabular}{|c|cc|ccc|}
% \hline
%       & \multicolumn{5}{c|}{FUNSD} \bigstrut\\
% \cline{2-6}Fusion method & mAP   & mRank & Hit@1 & Hit@2 & Hit@5 \bigstrut\\
% \hline
% \hline
% Concatenation & 0.7028 & 2.97  & 56.36 & 74.38 & 87.71 \bigstrut[t]\\
% Feature shifting w/o gate & 0.7024 & 3.10 & 55.82 & 74.76 & 88.84  \bigstrut[b]\\
% \hline
% \textbf{Feature shifting with gate (ours)} & \textbf{0.7177} & \textbf{2.89} & \textbf{58.19} & \textbf{76.27} & \textbf{88.94} \bigstrut\\
% \hline
% \end{tabular}%
%   \label{tab:fusion_b}%
% \end{table*}%

\subsection{Case Study}
\begin{figure}[h]
    \centering
    \includegraphics[width=\linewidth]{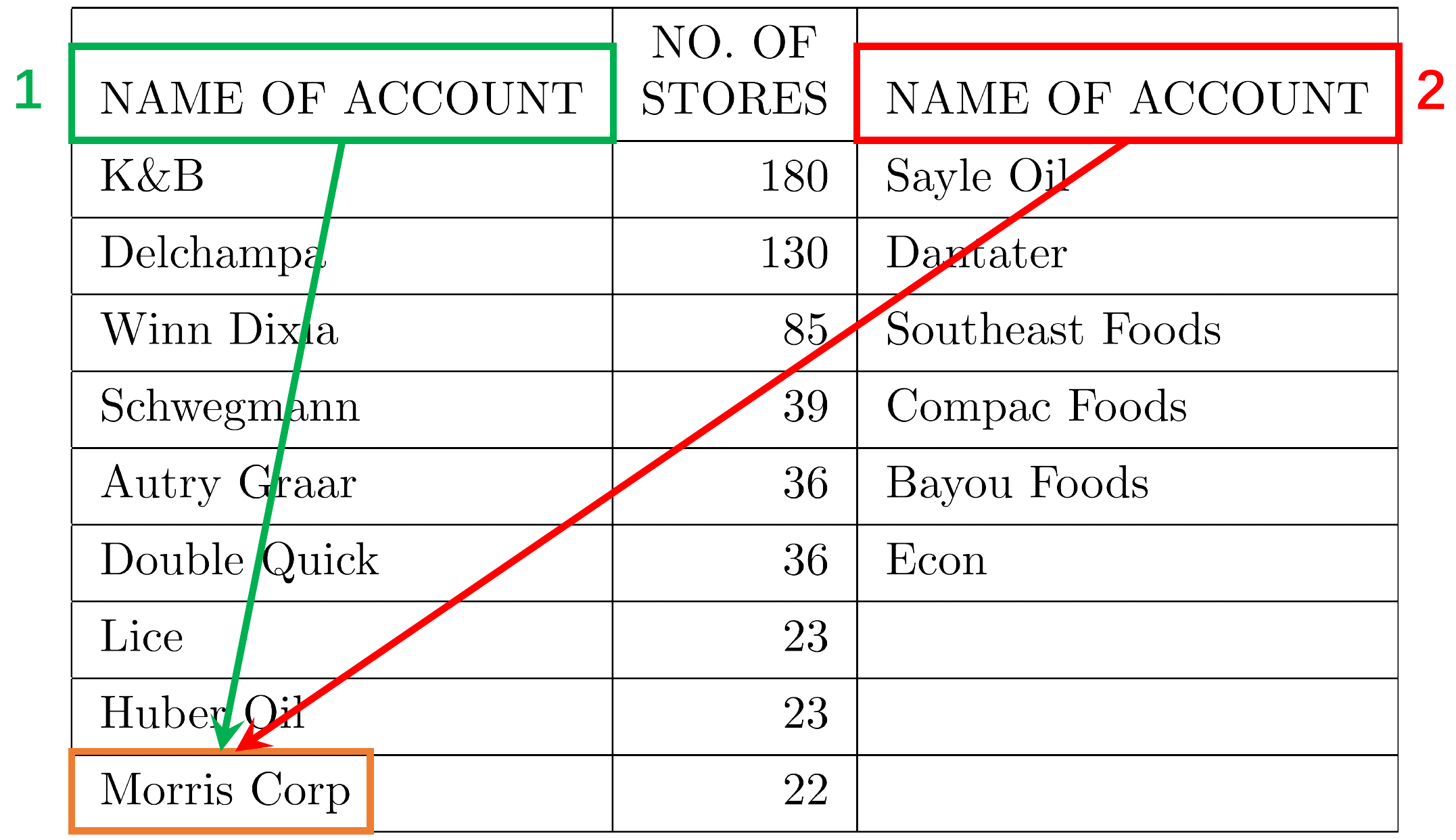}
    \caption{Case 1 from FUNSD: 
        the baseline of \{S, L\} predicts
        \textcolor[RGB]{0,176,80}{[NAME OF ACCOUNT (1)] $\to$}
        \textcolor[RGB]{234,112,13}{[Morris Corp]};
        the baseline of \{S\} predicts
        \textcolor{red}{[NAME OF ACCOUNT (2)] $\to$}
        \textcolor[RGB]{234,112,13}{[Morris Corp]}.}
    \label{fig:case_1}
\end{figure}

\begin{figure}[h]
    \centering
    \includegraphics[width=0.6\linewidth]{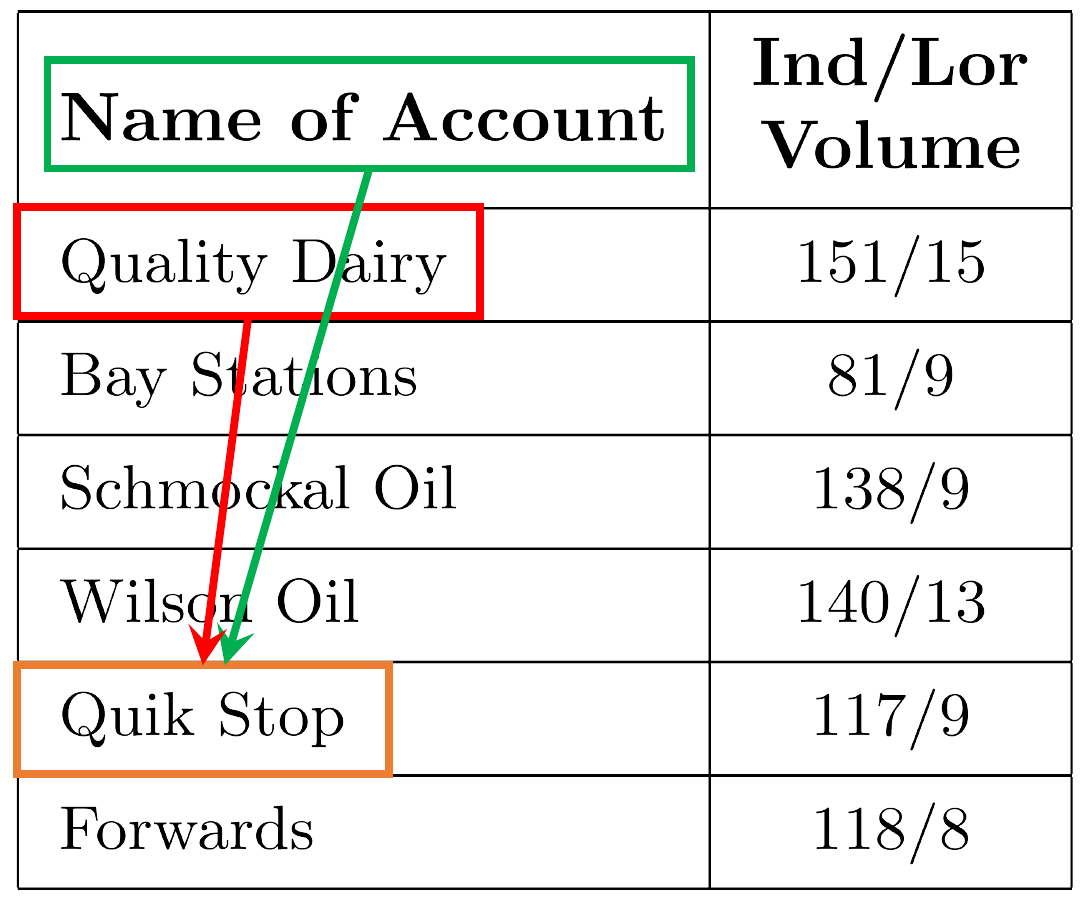}
    \caption{Case 2 from FUNSD: 
        our proposed model predicts \textcolor[RGB]{0,176,80}{[Name of Account] $\to$} \textcolor[RGB]{234,112,13}{[Quik Stop]};
        the baseline of \{S, L\} predicts 
        \textcolor{red}{[Quality Dairy] $\to$} \textcolor[RGB]{234,112,13}{[Quik Stop]}.}
    \label{fig:case_2}
\end{figure}

Figure \ref{fig:case_1} and \ref{fig:case_2} shows two examples from experiments on FUNSD dataset. Through the two real cases, the need of multimodal features has once again been proven.

In Figure \ref{fig:case_1}, we ask the baseline of \{S\} and the baseline of \{S, L\} to predict the superior counterpart for the text fragment [Morris Corp]. According to the label, its right answer is the text fragment [NAME OF ACCOUNT (1)] which is right above it. However, there are more than one fragments whose textual content is ``NAME OF ACCOUNT". The baseline of \{S\} cannot distinguish them and gives a wrong answer. 

 In Figure \ref{fig:case_2}, we compare the baseline of \{S, L\} with our proposed model. They predict the superior counterpart for the fragment [Quik Stop]. The right answer is [Name of Account], the topic at the head of the column. Although semantic contents and layout information are considered, the baseline of \{S, L\} cannot give the right answer. After adding the extra visual features and fusing them with the feature fusion module, our proposed model considers the bold and larger letters, and then gives the right answer. 

\subsection{Error Analysis}
We also observe some errors when our proposed method processes forms in our datasets. Although our proposed method has provided satisfactory result and can predict the right superior counterpart in most cases, further analysis of error cases is helpful to our future research.

\begin{figure}[h]
    \centering
    \includegraphics[width=0.9\linewidth]{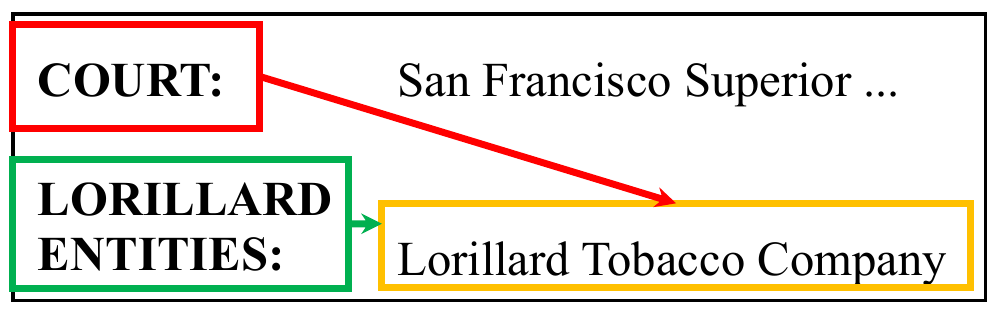}
    \caption{Error 1 from FUNSD: our proposed model wrongly predicts
    \textcolor{red}{[COURT:] $\to$} \textcolor[RGB]{234,112,13}{[Lorillard Tobacco Company]}. The probability is $0.9865$. The right answer is \textcolor[RGB]{0,176,80}{[LORRILLARD ENTITIES] $\to$} \textcolor[RGB]{234,112,13}{[Lorillard Tobacco Company]}.}
    \label{fig:error_1}
\end{figure}

In Figure \ref{fig:error_1}, our model make a mistake when predicting superior part of the fragment [Lorillard Tobacco Company]. The probability produced by our model is $0.9865$. The right answer's probability is $0.0107$ and ranks 2 among all 18 candidates. We attribute the error to the use of unknown word. \textit{"Lorillard"} is an uncommon word and the tokenizer will map all unknown words to the same token \verb|[UNK]|. The model cannot learn the relation without enough semantic information.

\begin{figure}[h]
    \centering
    \includegraphics[width=0.9\linewidth]{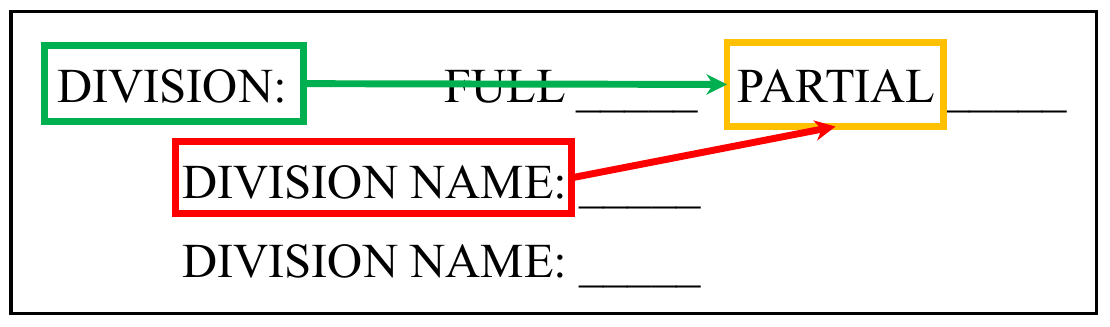}
    \caption{Error 2 from FUNSD: our proposed model wrongly predicts
    \textcolor{red}{[DIVISION NAME:] $\to$} \textcolor[RGB]{234,112,13}{[PARTIAL]}. The probability is $0.6686$. The right answer is \textcolor[RGB]{0,176,80}{[DIVISION] $\to$} \textcolor[RGB]{234,112,13}{[PARTIAL]}.}
    \label{fig:error_2}
\end{figure}

In Figure \ref{fig:error_2}, our model make a mistake when predicting superior part of the fragment [PARTIAL]. The probability produced by our model is $0.6686$. The right answer's probability is $0.3313$ and ranks 2 among all 75 candidates. We attribute the error to the too similar textual contents and nearer position. 

\section{Related Work}
\paragraph{Form Understanding}
Form understanding depends on two sub-tasks: the recognition of textual contents and the construction of the structure. Numerous existing works have produced satisfactory solutions for the first task \cite{liao2017textboxes,deng2018pixellink,liu2016star,wang2017gated}. Meanwhile, different directions have been proposed to deal with the form structure which is our focus in this paper.

Structure analysis used to be based on heuristic methods with handcrafted features \cite{ha1995recursive,simon1997fast,ha1995document}. These early jobs focus on the segmentation and layout of document pages, which provides coarse structural information. Some recent works adopt table detection techniques to build the structure \cite{hao2016table,he2017multi}. Table candidates are first selected out through some basic rules and then further filtered by convolutional neural network (CNN). In this way, the structure of forms can be clearly figured out when the contents are associated in a table. In our task, however, form is a more general concept and does not necessarily contain a table. It should be a collection of key-value pairs, which forms a hierarchical structure. We catch the basic components of a form document and design a reliable pipeline to extract the hierarchy through multiple modalities.

\paragraph{Feature Extraction}
We acquire the features through three modalities: semantic, layout, and vision. To extract semantic features from texts , the performance of pretrained language models, e.g. BERT \cite{devlin2018bert}, RoBERTa \cite{liu2019roberta}, has been proved in many natural language processing tasks. And these models can also be used in many downstream tasks through fine-tuning. As for visual feature extraction, methods based on convolutional neural networks are widely studied and used \cite{he2016deep,simonyan2014very,szegedy2015going}. A recurrent neural network layer is usually added to further suit the character sequence \cite{shi2016end}. We follow these previous works and design the feature extraction modules in our model.

\paragraph{Feature Fusion}
Multimodal features is commonly used to improve performance. Many fusion methods are proposed to properly utilize the features. In \citet{poria2017context}, direct concatenation is used to get a joint representation for a sentence. A two-stage fusion hierarchy is proposed as well \cite{majumder2018multimodal}. Previous works have also proposed a feature shifting method to use shifting feature to fix the base features \cite{wang2019words}.

\section{Conclusion}
In this paper, we proposed a multimodal method to extract key-value pairs and build the hierarchy in forms to improve the general form understanding. We leveraged advanced models, e.g. BERT, Resnet, LSTM, to acquire features from multiple aspects: semantic, layout and vision. For the first time, heterogeneous features are combined to extract the hierarchical structure in forms. And the proposed hybrid fusion method of concatenation and feature shifting effectively obtains the joint feature and eliminates the interference. We also adopted negative sampling technique to train our model. Furthermore, extensive experiments demonstrate the advantages of our method to build the form hierarchy.

In the future, we will apply our method to other challenging benchmarks and strive to combine the idea of pretraining with our method.

\bibliography{ref}
\bibliographystyle{acl_natbib}

\clearpage

\section*{Appendix}

\appendix 

\section{Details about Datasets}
\subsection{MedForm}
MedForm dataset is built by us. Our research group cooperates with medical examination agencies and collects a large number of medical examination forms. The forms may be from different agencies, so the formats are not the same. We recruit 10 skilled annotators to label the data and another 5 annotators to further check the quality. The annotation includes the area of text fragments in a form page, the textual contents in each fragment, the exact coordinates and the hierarchical relation between the fragments. Because of privacy concerns, we cannot make this dataset public.

\subsection{FUNSD}
FUNSD dataset is a new and public dataset for form understanding tasks. It consists of various real fully annotated, scanned forms. It offers the textual contents and exact coordinates of each fragments. The dataset can be downloaded in \url{https://guillaumejaume.github.io/FUNSD/}.

% \section{Original Example Images}
% The original scanned forms in FUNSD dataset are not clear. To make the paper more readable, we use the processed and translated images in the paper. Original images can be found here.
% \begin{figure}[h]
%     \centering
%     \includegraphics[width=0.8\linewidth]{fig/structure.pdf}
%     \caption{Original Image of Figure \ref{fig:example}.}
% \end{figure}

% \begin{figure}[h]
%     \centering
%     \includegraphics[width=\linewidth]{fig/example_1.pdf}
%     \caption{Original Image of Figure \ref{fig:case_1}.}
% \end{figure}

% \begin{figure}[h]
%     \centering
%     \includegraphics[width=0.6\linewidth]{fig/example_2.pdf}
%     \caption{Original Image of Figure \ref{fig:case_2}.}
% \end{figure}

\section{Metric Explanation}
\subsection{Mean Average Precision (mAP)}
Mean Average Precision is a metric widely used in the area of object detection. It measures the average precision value for different recall value, so the larger mAP is, the better the model performs. 

In our reconstruction task, we detect the superior counterpart for a given text fragment. For example, the given text fragment $x$ has $n$ candidates $y_1,y_2,...,y_n$ and $m$ of them are the right answers. The recall value can be $\frac{i}{m}$, where $i = 0,1,...,m$. We calculate the biggest precision value for each recall value. We denote the biggest precision when recall equals $\frac{i}{m}$ as $p_i$ and calculate mAP as followed:
\begin{equation}
    mAP = \sum_{i=0}^m p_i * \frac{1}{m}
\end{equation}

\subsection{Mean Rank (mRank)}
Mean Rank is also a metric used in the area of object detection. It measures the average number of wrong answers that rank higher than right answers, so the smaller mRank is, the better the model performs. 

It is easier to think the mRank as the average number of right-wrong reverse pairs in ranking list. For example, our given text fragment $x$ has $n$ candidates $y_1,y_2,...,y_n$. Among the candidates, $m$ of them are right answers and the corresponding indices are $i_1,i_2,...,i_m$ in an ascending order. For the first right answer $y_{i_1}$, the number of wrong answers that rank higher than it is $i_1 - 1$. For the second right answer $y_{i_2}$, the number of wrong answers that rank higher than it is $i_2 - 2$ (excluding $y_{i_1}$ before it). Therefore, we calculate mRank as followed:
\begin{equation}
    mRank = \sum_{k=1}^m i_k - k = \sum_{k=1}^m i_k - \frac{(1+m)m}{2}
\end{equation}

\subsection{Hit@k}
Hit@k is a metric that measures the ratio of cases whose right answers appear in the top $k$ prediction candidates. In our detection task, we detect the superior counterpart for a given text fragment. We predict the probability for every text fragment in a page and count the number of fragments whose true superior counterparts appears in the top $k$ candidates. We calculate Hit@k through division of the number and the total number of fragments. 
	
% \begin{figure*}[!h]
% 	\centering
% 	\subfigure[]{
% 		\includegraphics[width=0.3\linewidth]{fig/structure.pdf}
% 	}
% 	\subfigure[]{
% 		\includegraphics[width=0.3\linewidth]{fig/example_1.pdf}
% 	}
% 	\subfigure[]{
% 		\includegraphics[width=0.3\linewidth]{fig/example_2.pdf}
% 	}
% 	\caption{Original Example Images of Figure \ref{fig:example}, \ref{fig:case_1}, \ref{fig:case_2}.}
% \end{figure*}

\end{document}